\documentclass[letterpaper, 10 pt, conference]{ieeeconf}  % Comment this line out if you need a4paper
\pdfoutput=1
\IEEEoverridecommandlockouts                              % This command is only needed if 
\usepackage{graphics} % for pdf, bitmapped graphics files
\usepackage{epsfig} % for postscript graphics files
\usepackage{times} % assumes new font selection scheme installed
\usepackage{amsmath} % assumes amsmath package installed
\usepackage{amssymb}  % assumes amsmath package installed
\usepackage{multirow}
\usepackage{array}
\usepackage{booktabs}
\usepackage{caption}
\usepackage{subfigure}
\usepackage{latexsym}
\usepackage{dsfont}

\title{\LARGE \bf
VisionNet: A Drivable-space-based Interactive Motion Prediction Network for Autonomous Driving 
}

\author{Yanliang Zhu, Deheng Qian, Dongchun Ren, Huaxia Xia% <-this % stops a space
\thanks{*This work was supported by the Meituan-Dianping Group.}% <-this % stops a space
\thanks{Yanliang Zhu, Deheng Qian, Dongchun Ren and Huaxia Xia are with the Meituan-Dianping Group, Beijing, China.
       {\tt\small zhuyanliang@meituan.com}}%
}

\begin{document}

\maketitle
\thispagestyle{empty} %forces first page to not have a header
%\pagestyle{empty}

%\ieeefootline{Workshop on Latex Style Files \\ International Conference on Latex 2014, Las Vegas, NV, USA}%creates footline

%\ieeeheadline{Workshop on Latex Style Files \\ International Conference on Latex 2014, Las Vegas, NV, USA}%creates headline

%%%%%%%%%%%%%%%%%%%%%%%%%%%%%%%%%%%%%%%%%%%%%%%%%%%%%%%%%%%%%%%%%%%%%%%%%%%%%%%%
\begin{abstract}
The comprehension of environmental traffic situation largely ensures the driving safety of autonomous vehicles.
Recently, the mission has been investigated by plenty of researches, while it is hard to be well addressed due to the limitation of collective influence in the complex scenarios.
These approaches model the interactions through the spatial relations between the target obstacle and its neighbors.
However, they oversimplify the challenge since the training stage of the interactions lacks effective supervision. 
As a result, these models are far from promising.
More intuitively, we transform the problem into calculating the interaction-aware drivable spaces and propose the CNN-based VisionNet for trajectory prediction.
The VisionNet accepts a sequence of motion states, i.e., location, velocity and acceleration, to estimate the future drivable spaces.
The reified interactions significantly increase the interpretation ability of the VisionNet and refine the prediction.
To further advance the performance, we propose an interactive loss to guide the generation of the drivable spaces.
Experiments on multiple public datasets demonstrate the effectiveness of the proposed VisionNet.
\end{abstract}

%%%%%%%%%%%%%%%%%%%%%%%%%%%%%%%%%%%%%%%%%%%%%%%%%%%%%%%%%%%%%%%%%%%%%%%%%%%%%%%%
\section{INTRODUCTION}

Predicting future trajectories of surrounding dynamic traffic agents (e.g., vehicles, bicycles and pedestrians) is an important technology to build a self-driving vehicle. The prediction helps the self-driving vehicle plan a safe and smooth trajectory. However, it is a challenge for the vehicle to make proper predictions in complex traffic scenarios.

Specifically, predicting future trajectories of other agents suffers from two challenges. First, the observed historical trajectories of agents are noisy, which is inevitable in data measurement and processing. Second, the trajectory of each agent is influenced by complex interactions among the surrounding agents. For example, a traffic agent plans its trajectory according to both its intention and the environment. To handle such two challenges, there exist multiple algorithms which can be classified into two categories, i.e., coordinate-based prediction algorithms \cite{c4,c5,c6} and vision-based prediction algorithms \cite{c1,c2,c3}. 

The coordinate-based algorithms leverage the measurement data, including location, acceleration and orientation information, then forecast the future positions in the world or ego-vehicle coordinates. They build submodules like social grid \cite{c12}, interaction layer \cite{c7} and world model \cite{c13} to capture the interactions over all obstacles. However, such methods lack the supervision for learning the interactions. 
\begin{figure}[thpb]
	\centering
	\includegraphics[width=8.5cm]{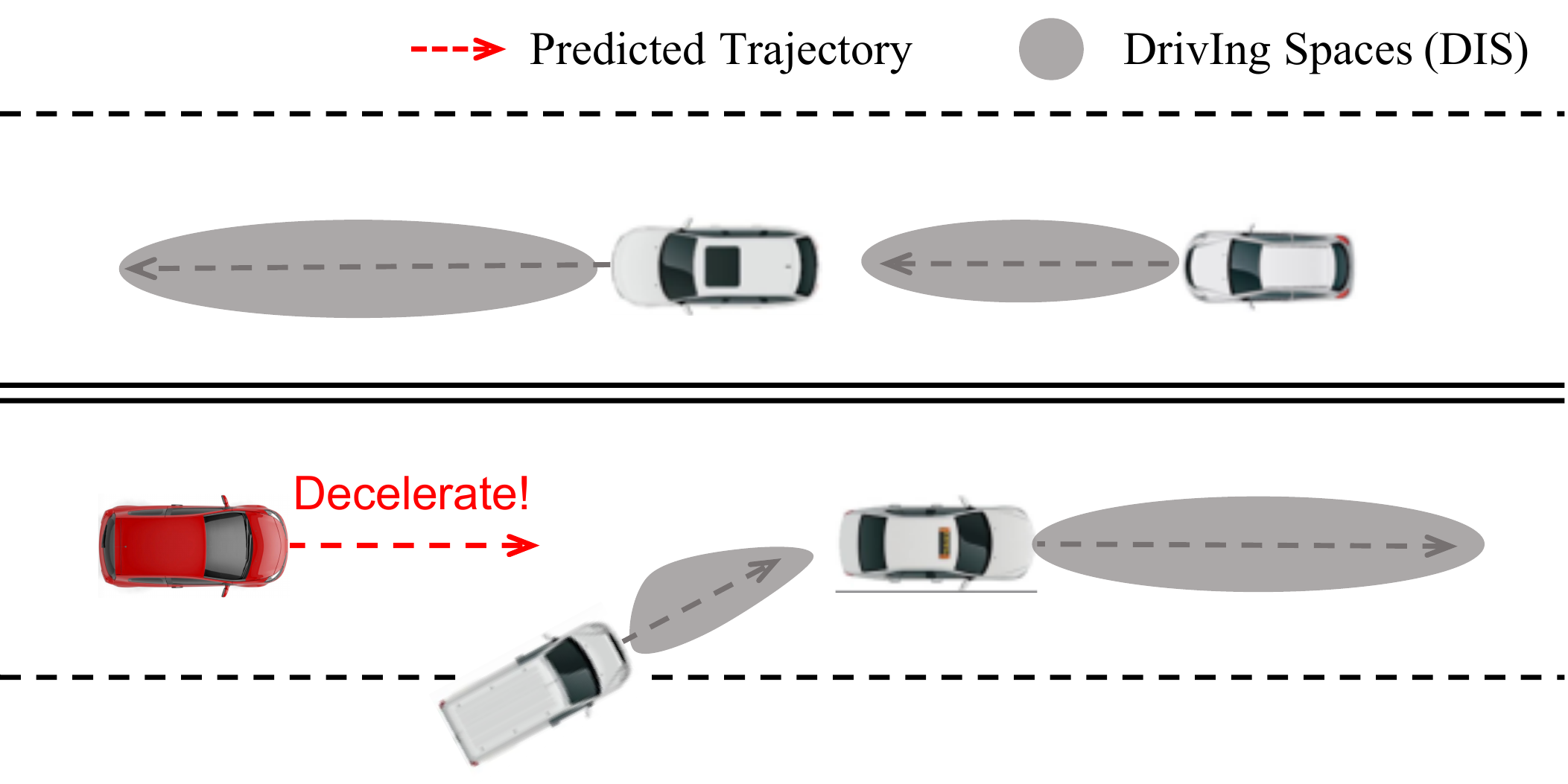}
	\caption{
		Illustration of the motion prediction task. Having a predicted vehicle with red color, the autonomous driving system estimates the driving spaces of all the surrounding obstacles and infers the interaction-aware global drivable spaces. The resulting drivable spaces are used to assist in trajectory generation.
	}
	\label{fig:1}
\end{figure}
As a result,  they can hardly present the collective influence precisely, leading to inaccurate predictions. By contrast, the vision-based algorithms are capable of processing semantic ``images'' like camera images, LiDAR point clouds and Occupancy Grid Maps (OGMs). 
Works by \cite{c1,c2} generate image sequences using Convolutional Neural Networks (CNNs). These approaches demonstrate a strong understanding of the environment,  however, they neglect the complex interactions among traffic agents. Particularly, existing methods do not deal with measurement and process noises in the input data.

In this paper, we propose an interaction module into a vision-based architecture for predicting trajectories of surrounding obstacles effectively. Our approach, VisionNet, describes the interactions by calculating drivable spaces in the scene, which allows our model to be more interpretable. Specifically, the VisionNet consists of two networks, i.e., the Interaction Network and the Prediction Network. The Interaction Network adopts obstacles' motion states to infer global drivable spaces in the future. Towards this goal, the driving spaces for observed obstacles are calculated and then be used to determine the drivable spaces. The resulting drivable spaces possess higher energies in those areas where it is more likely to have a collision. Depending on the drivable spaces, as well as the observed trajectory, the Prediction Network predicts future trajectory. Besides, inspired by the fact that obstacles tend to move in safe regions, we introduce an interactive loss to supervise the training of drivable spaces. Specifically, given ground-truth positions, we minimize the energies at these positions in the predicted drivable spaces. 

In summary, our main contributions contain:

\begin{itemize}
	\item We propose to describe the interactive effects among obstacles with drivable spaces, which is more comprehensible and interpretable. 
	\item We propose to construct a more accurate framework to learn the interactions, as well as to predict future OGMs.
	\item We perform multiple experiments on the KITTI-Tracking \cite{c14} dataset, UCY \cite{c15} and ETH \cite{c16} pedestrian trajectories datasets, which demonstrates that VisionNet outperforms state-of-the-art methods in trajectory prediction.
\end{itemize}

\section{RELATED WORK}\label{section:2}

\subsection{Prediction Models}

Previous works present a motion prediction model with traditional machine learning techniques, including Support Vector Machines\cite{c8}, Gaussian Mixture Models\cite{c9}. 
However, they are extremely limited by the model capacity.
Recently, Neural Networks \cite{c10,c11} have proven to be effective in obstacle trajectory prediction. 
The prediction methods based on Neural Networks can be categorized into two groups, i.e., coordinate-based models and vision-based models.

The coordinate-based principles are widely practiced in conventional autonomous driving systems.
They accept the measurement data without any transformation and then perform a forward process.
In \cite{c5}, the authors proposed an approach based on Long Short Term Memory (LSTM).
The model extracted hand-crafted features such as displacement, angular changes histogram and orientation, then predicted motion behaviors of the dynamic objects.
Deo et al. \cite{c6} devised a double-LSTM to learn the maneuver and the path jointly.
Ju et al. \cite{c7} combined Kalman filtering with Neural Network and introduced the Interaction-aware Kalman Neural Networks (IaKNN) to predict the interaction-aware trajectories.
Moreover, a kinematic model was utilized to constrain the generated trajectories and refine forecasting.
Other researches \cite{c17,c18,c19,c20,c21} further extended the models to the pedestrian trajectory prediction in the free spaces.

When it comes to the vision-based prediction, tasks such as video-to-video synthesis \cite{c23}, image-to-video translation \cite{c3} and object tracking \cite{c24} have achieved decent performance. 
This motivates us to extend the generation models to obstacle motion prediction.
Srikanth et al. \cite{c1} took the raw camera and LiDAR data as input and consequently obtained context representations such as semantic, instance and depth maps.
These maps were then employed to predict the target trajectories.
The approach by \cite{c2} embedded the locations of the obstacles into OGMs for each past moment.
The encoder-decoder architecture with Convolutional-LSTM then generated new OGMs to produce the probability distribution of future positions.
In \cite{c22}, authors employed a probabilistic forecasting model called Estimating Social-forecast Probability (ESP).
With LiDAR point clouds and position coordinates, the ESP reasoned about how dynamic obstacles would be likely to move under the mutual influence.

Compared with coordinate-based approaches, the vision-based models usually acquire semantic information which is more comprehensive.
However, interaction modeling is still far from explainable since existing works lack supervision for interactions during the training process. 
In contrast, we introduce an interpretable interaction module to address this issue effectively.

\subsection{Interaction Modeling}

Actually, an obstacle's motion depends on both its intension and the surrounding environment.
To anticipate the future trajectories of the dynamic obstacles, interactions among traffic participants need to be clarified precisely.
Many types of researches aimed at capturing collective influence with respect to spatial relations. 
\cite{c5} trained a deep neural network with extracted features, including relative distance and velocity of nearby vehicles in the lateral and longitudinal directions.
Hu et at. \cite{c4} pointed out that the prediction of possible destinations could bring about more intuition.
Given certain targets with human prior knowledge, probability computation was conducted according to the observed path and its neighbors.
Ultimately, the optimal trajectory was hinted by the destination of the highest confidence.
Approach in \cite{c6} embedded relative positions of the main vehicle and its adjacent vehicles into a context vector. 
The context vector was then sent to the LSTM-based encoder-decoder architecture to assist in trajectory generation.
Alahi et al. \cite{c12} built a social pooling layer which divided the neighboring district into mesh grids and then embedded the pedestrians in the grid.

Above all, the investigations were mainly concentrated on the employment of relative positions.
However, the whole routine oversimplifies the interactions among dynamic obstacles, which counts a lot, especially in complex traffic scenarios.
Besides, it lacks interpretability as well as the supervision for interactions during the entire learning phase.
In this paper, our VisionNet views the interactions as drivable spaces. Meanwhile, an interactive loss is employed to improve the quality of the interactions.
with the engine above, the proposed network is capable of modeling mutual interactions overall agents, which is more elaborated and explainable.

\section{APPROACH}\label{section:3}

\begin{figure*}[thpb]
\centering
\includegraphics[width=17.6cm]{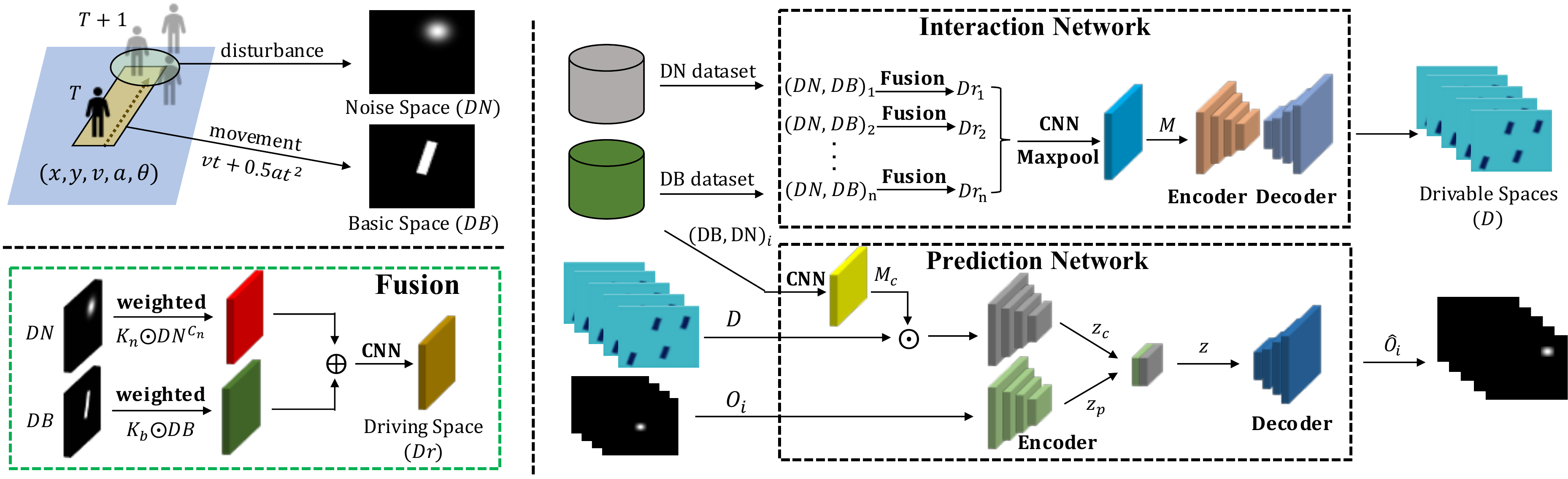}
\caption{
Main architecture of the VisionNet. The VisionNet involves an interaction network and a prediction network. The interaction network takes the basic driving spaces $DB$ and the normalized noise driving spaces $DN$ of all the obstacles and then infers the global drivable spaces $D$ in the future. The prediction network, take the $i$-th obstacle as an example, forecasts trajectory $\widehat{O}_i$ at the next moments using the observed trajectory $O_i$ and the above global drivable spaces $D$. Besides, $DB_i$ and $DN_i$ are leveraged to remove the ``self-predictive'' information in the global drivable spaces at the prediction stage.
}
\label{fig:2}
\end{figure*}

\subsection{Problem Formulation}
Occupancy Grid Maps (OGMs) divide a realistic area into the mesh grids, where each grid indicates the observation probability of the target object. The value of the grid located at the $i$-th row and the $j$-th column in an OGM is denoted as $o\left(i, j\right) \in \left[0, 1\right]$. Once a mapping relation between a road area of $R \times C$ square meters and an OGM of $H \times W$ square pixels is determined, the mathematical formulas between the location $\left(x, y\right)$ in the area and the location $\left(h, w\right)$ in the OGM can be represented as,
\begin{align*}
&h = \left(H / R\right)  x, \  w = \left(W / C\right)  y \tag{1} \\
& x = \left(R / H\right)  \left(h + 0.5 \right), \ y = \left(C / W\right)  \left(w + 0.5\right) \tag{2}
\end{align*}

Assume the number of the observed obstacles is $N$. The observation time steps and the total time steps are $T_{obs}$ and $T_{pred}$ respectively. 
In other words, the mission is to predict future positions between $T_{obs}+1$ and $T_{pred}$ according to the previous positions of all the obstacles from the first time step to $T_{obs}$.
For the $i$-th individual, $X_{i} \in \mathbb{R}^{T_{obs}\times5}$ and $Y_{i} \in \mathbb{R}^{T_{obs}\times5}$ stand for its historical and future groundtruth trajectories respectively.
Each line in the $X_i$ and $Y_i$ matrixes is a 5-tuple $(x, y, v, a, \theta)$.
$\left(x, y\right)$ is an obstacle's position, $\theta$ is the moving orientation, $v$ denotes the speed and $a$ represents the acceleration.
The prediction issue is then fomulated as searching for the optimal mapping function from $\mathbb{X}=\left\{X_{i}\right\}_{i=1}^N$ to $\mathbb{Y}=\left\{Y_{i}\right\}_{i=1}^N$.
In addiction, we denote the predicted trajectory as $\widehat{Y}_{i}$.
According to Eq. (1), the trajectory $X_i$ (or $Y_i$) can be easily transformed into the OGMs $O_i \in \mathbb{R}^{T_{obs} \times H \times W}$ (or $O_{gt,i} \in \mathbb{R}^{\left(T_{pred}-T_{obs}\right) \times H \times W}$) by the computation.

As shown in Figure \ref{fig:2}, the VisionNet is proposed to estimate the prediction function. It is composed of two deep networks, the interaction network and the prediction network.
The interaction network is fed by the past trajectories of all obstacles and computes the Global DrivAble Spaces (GDAS) from $T_{obs+1}$ to $T_{pred}$.
Actually, the GDAS represents a comprehensive influence among the traffic agents.
Then, depending on the historical trajectory and the GDAS, the prediction network produces the future trajectory for each obstacle.
Different from conventional coordinate-based methods, our approach models the global interactive effects among obstacles and takes advantage of OGMs for future trajectory prediction.
The VisionNet can consequently take in interactions and outperform previous methods in both complex conditions and normal scenarios.

\subsection{The Interaction Network}
Under the collective influence of all obstacles, the interaction network aims at predicting the GDAS in the future $D \in \mathbb{R}^{\left(T_{pred}-T_{obs}\right) \times H \times W}$. 
For example, given OGMs $\mathbb{O}=\left\{O_{i}^{t}\right\}_{i=1}^N$ of $N$ static obstacles at time step $t$, the value at the location $\left(x,y\right)$ in the drivable spaces at the next moment could be expressed by the probability formula,
$$
D^{t+1}\left(x,y\right) = 1 - \prod_{i=1}^{N} \left(1-O_i^t\left(x,y\right)\right) \eqno{(3)}
$$

Nevertheless, most traffic participants behave dynamically throughout rather than staying still.
An intuitive solution is to predict DrivIng Spaces (DIS) for each obstacle  first and then generates GDAS based on individual results.
Accordingly, the interaction network can be written as,
\begin{align*}
& D = \mathsf{f}\left(\mathsf{g}\left(X^1\right),\mathsf{g}\left(X^2\right), \cdots, \mathsf{g}\left(X^{T_{obs}}\right)\right) \tag{4} 
\end{align*}
where $\mathsf{g}\left( \cdot \right)$ maps the past discrete positions to the DIS.
The integration function $\mathsf{f}\left( \cdot \right)$ fuses all the DIS and estimates GDAS.
These two functions are implemented as convolutional layers in deep neural networks.
Next, detailed descriptions of the two mapping networks will be introduced respectively.

\subsubsection{Positions to Driving Spaces}
In a dynamic environment, both interactions and data noise challenge the computation of drivable spaces.
Because of the measurement precision and data-processing mechanism, noise and delay are likely to appear in observed trajectories.

For example, in the data processing phase, an autonomous driving system generally employs a difference method to determine obstacle's motion states (e.g. velocity, acceleration and orientation).
The formulas could be expressed as follows,
\begin{align*}
& v_i^t = \left(p_i^t - p_i^{t-1}\right) / \Delta t \tag{5} \\
& a_i^t = \left(p_i^t + p_i^{t-2} - 2p_i^{t-1}\right) / \Delta t^2 \tag{6}\\
& \theta_i^t = \mathsf{arctan}(p_i^t, p_i^{t-1}) \tag{7}
\end{align*}
where $p_i^t$ is the world/vehicle coordinate $\left(x_i^t, y_i^t\right)$ of the $i$-th obstacle at time step t.
$\Delta t$ is the time interval between two adjacent moments.
Function $\mathsf{arctan}\left(\cdot\right)$ is to calculate the orientation.
Since the motion states would always include a time delay of $\Delta t$.
The deviation of observed trajectories caused by the delay becomes more significant in high-speed environment.

In this part, we take into account the disturbance when generating DIS.
To address the problem, we propose a new mathematical model to produce driving areas accurately.
Here, like Kalman filtering \cite{c7}, we correct the existing motion states using Gaussian random noise,
\begin{align*}
& \widehat{a}_i^t = a_i^t + \omega_i^t + \xi_i^t\tag{8}\\
& \widehat{v}_i^t = v_i^t  + \omega_i^t   \Delta t +\varphi_i^t \tag{9} \\
& \widehat{\theta}_i^t = \theta_i^t + \eta_i^t +\psi_i^t \tag{10}
\end{align*}
where $\omega_i^t \sim \mathcal{N} \left(0,Q_i^t \right) $ and $\eta_i^t \sim \mathcal {N} \left(0,R_i^t \right)$ are Gaussian noises with zero mean and specific variance for the $i$-th obstacle respectively.
These two terms are used to correct the process deviation.
Variables $\xi_i^t \sim \mathcal N \left(0,S_i^t \right) $, $\varphi_i^t \sim \mathcal {N} \left(0,U_i^t \right)$ and $\psi_i^t \sim \mathcal N \left(0,V_i^t \right) $ are Gaussian noises which describe the measurement noises during the data collection. Here,  $Q_i^t$, $R_i^t$, $S_i^t$, $U_i^t$ and $V_i^t$ are the different variances.
According to classical kinematic models, the traveled distance from the present moment to the next can be written as,
\begin{align*}
s_i^t &= \widehat{v}_i^t  \Delta t + \frac{1 }{2} \widehat{a}_i^{t}  \Delta t^2 \\
&=v_i^t  \Delta t +  \frac{1 }{2}  a_i^{t}  \Delta t^2 + \left(\omega_i^t   \Delta t^2 +  \frac{1 }{2}  \omega_i^{t}  \Delta t^2\right) \\
&\ \ \ + \left(\varphi_i^t  \Delta t +  \frac{1 }{2}  \xi_i^t  \Delta t^2\right) \tag{11}\\
&=v_i^t  \Delta t +  \frac{1 }{2}  a_i^{t}  \Delta t^2  + \left( \frac{3}{2}  \omega_i^{t}  \Delta t^2 +\varphi_i^t  \Delta t +  \frac{1 }{2}  \xi_i^t  \Delta t^2\right)
\end{align*}

As $\omega_i^t$, $\xi_i^t$ and $\varphi_i^t$ are independent variables, $s_i^t$ still obeys the Gaussian distribution as follow,
\begin{align*}
&s_i^t \sim \mathcal{N} \left(\mu_i^t, P_i^t \right) \\
&\mu_i^t = v_i^t  \Delta t +  \frac{1 }{2}  a_i^{t}  \Delta t^2 \tag{12}\\
&P_i^t =  \frac{9}{4}  \Delta t^4  Q_i^t + \Delta t^2  U_i^t +  \frac{1 }{4}  \Delta t^4  S_i^t 
\end{align*}

Then, the DIS can be obtained with the corrected traveled distance $s_i^t$.
More specifically, we divide the DIS into two parts, the Basic DrivIng Spaces (B-DIS) and the Noise DrivIng Spaces (N-DIS).
B-DIS represents the fundamental moving region controlled by the measured velocity and acceleration.
The length of the moving region is $\mu_i^t$, and the width of the moving region is the same as the width of the obstacle.
The orientation of the region is $\theta_i^t$. 
N-DIS captures the disturbance which is dominated by both the measurement noise and the process noise.
The center of the noise area in N-DIS falls in location $\left(x_i^t+\mu_i^t  cos\left(\theta_i^t\right), y_i^t + \mu_i^t  sin\left(\theta_i^t\right)\right)$.
The variance of the noise area is $P_i^t$. 
In fact, it is difficult to determine the variance matrix $P_i^t$ which depends on an obstacle's movement in the past.
In this paper, we employ a convolutional neural network to learn the variance using the obstacle's motion state. Specifically, The grid value in N-DIS can be transformed as follows, 
\begin{align*}
NDIS\left(x,y\right) \sim \mathcal{N} \left(\mu_i^t, P_i^t \right) \varpropto \mathcal{N} \left(\mu_i^t, 1 \right) ^ {1 / P_i^t} \tag{13}
\end{align*}
where $NDIS\left(x,y\right)$ is the value of position $\left(x,y\right)$ in N-DIS.
We denote the B-DIS as $DB$ and the normalized N-DIS as $DN$ which follows the Gaussian distribution $\mathcal{N}\left(\mu_i^t, 1 \right)$.
These two types of spaces are taken as input of VisionNet.

Finally, our network for producing DIS can be written as,
\begin{align*}
Dr_i^t &=\mathsf{g}\left(X_i^t\right) \\
&= \mathsf{h}\left( \mathsf{K_b} \left(X_i^t\right) \odot DB_i^t + \mathsf{K_n} \left(X_i^t\right) \odot NDIS_i^t \right)  \tag{14} \\
&= \mathsf{h}\left( \mathsf{K_b} \left(X_i^t\right) \odot DB_i^t + \mathsf{K'_n} \left(X_i^t\right) \odot {DN_i^t}^ {\mathsf{C_n}\left(X_i^t\right)} \right)
\end{align*}
where $\mathsf{K_b}\left(\cdot\right)$ and $\mathsf{K'_n}\left(\cdot\right)$ compute weight matrixes of B-DIS and N-DIS respectively. $\mathsf{C_n}\left(\cdot\right)$ computes the variance matrix $P_i^t$.
The functions $\mathsf{K_b}$, $\mathsf{K'_n}$ and $\mathsf{C_n}$ are achieved by convolutional neural networks. $\odot$ denotes element-wise multiplication and $\mathsf{h}\left(\cdot\right)$ is a convolutional layer to capture disturbance of the orientation. $Dr_i^t$ is the driving spaces for the $i$-th obstacle at time step $t$.

\subsubsection{Integration of Driving Spaces}
The VisionNet simultaneously takes the DIS of all obstacles in the past and forecasts the future GDAS.
The output of the prediction network represents drivable spaces under the collective influence of the traffic participants.
The operation formula of the prediction network is written as,
\begin{align*}
\mathbf{f}: \{Dr_i\}_{i=1}^N \longmapsto D \tag{15}
%\mathbb{D} = \mathsf{f}\left(D_1, D_2, \cdots, D_N \right) \tag{16}
\end{align*}
where $Dr_i  \in \mathbb{R}^{T_{obs} \times H \times W}$ is the sequential driving spaces integrated by the aforementioned B-DIS and N-DIS for the $i$-th individual.
Next, we will provide the details about this mapping relations.

In the encoding phase, we use convolutional and maxpool functions to generate global drivable spaces.
Maxpool is widely used for processing variable and unordered data \cite{c13,c27}.
The equations of the encoder can be written as,
\begin{align*}
&M^t = \mathsf{Maxpool}\left(\mathsf{C_p}\left(Dr_1^t\right), \cdots, \mathsf{C_p}\left(Dr_N^t\right) \right)  \tag{16} \\
&z_k = \mathsf{Encode_k}\left(M^1, M^2, \cdots, M^{T_{obs}}\right) \tag{17}
\end{align*}
where $\mathsf{C_p}\left(\cdot\right)$ is a convolutional function.
$M^t$ is a fusion map that represents the global driving spaces overall dynamic obstacles at time step $t$. 
$\mathsf{Encode_k}\left(\cdot\right)$ is the encoding function with $T_{obs}$ input channels.
$z_k$ is the embedding feature and it will be sent to decoder to predict global drivable areas in the future.
\begin{align*}
D  = \mathsf{Decode_k}\left(z_k\right) \tag{18}
\end{align*}
where $\mathsf{Decode_k}\left(\cdot\right)$ is the decoding function which composes of multiple deconvolutional layers.
 
\subsection{The Prediction Network}
In this section, based on the observed OGMs $\mathbb{O}$ and the interaction-aware drivable spaces $D$, we present how to establish a CNN-based architecture to forecast obstacles' future OGMs $\mathbb{\widehat{O}}$.
As shown in Figure \ref{fig:2}, the prediction network contains two parts, i.e., feature fusion and image synthesis.

\subsubsection{Feature Fusion}
We firstly use an encoder to embed the observed OGMs into a latent feature map $z_p$. 
As for drivable spaces, we need to remove the ``self-predictive'' information.
Specifically, the GDAS indicates safety and collision regions in the scene.
Those collision areas are determined by the movement of all the obstacles, including the predicted obstacle.
In other words, the GDAS contains ``self-predictive'' information.
If the GDAS was directly adopted to predict the future motion of the obstacle, it could confuse the prediction network and lead to a poor performance. 

To address the problem, we leverage an ablation mask to filter out the ``self-predictive'' regions in GDAS.
Meanwhile, an additional encoder is employed to transform the filtered drivable spaces into a latent feature map $z_c$.
The $z_p$ and $z_c$ are then concatenated as a fusion feature map $z$, which is sent to the decoder to predict future images.
The formulas of the feature fusion can be expressed as,
 \begin{align*}
&M_c = \mathsf{f_c}\left(1-DB\right) + \mathsf{f_c}\left(1-DN\right) \tag{19} \\
&z = \mathsf{Concat}\left(\mathsf{Encode_p}\left(\mathbb{O}\right), \mathsf{Encode_c}\left(M_c \odot D \right)\right) \tag{20}
\end{align*}
where $\mathsf{f_c}\left(\cdot\right)$ is the mask generation function with a simple convolutional structure. 
Particularly, purely for convenience, we use the normalized N-DIS instead of the completed N-DIS in above function.
$\mathsf{Encode_c}\left(\cdot\right)$ and $\mathsf{Encode_p}\left(\cdot\right)$ are the encoders for processing observed OGMs and filtered GDAS.

\subsubsection{Image Synthesis}
With the generated fusion feature map $z$, there are alternative techniques to produce OMGs as sequential issues, such as Conv-LSTM \cite{c2} and CNN \cite{c3}.
Similar to \cite{c3}, we employ several upsampling deconvolutional layers to synthesize images.
The output of the decoder is defined as follow,
\begin{align*}
&\mathbb{\widehat{O}} = \mathsf{Decode_p}\left(z\right) \tag{21}
\end{align*}

\subsection{Implementation Details}
In this section, elaborated architecture of the VisionNet, as well as the loss function will be introduced. 
$\left(K, S, C\right)$ represents the configuration of the convolutional or deconvolutional layer. $K$ and $S$ are the kernel size and stride respectively while $C$ stands for the output channels.

\subsubsection{Network Configuration}
In the interaction network, each $\mathsf{K_b}\left(\cdot\right)$, $\mathsf{K'_n}\left(\cdot\right)$ or $\mathsf{C_n}\left(\cdot\right)$ function consists of three layers, i.e., a fully connected layer with $T_{obs}WH/8$ output neurons,  two deconvolutional layers $\mathsf{D1}\left(4, 2, T_{obs}/2\right)$ and  $\mathsf{D2}\left(4, 2, T_{obs}\right)$. 
$\mathsf{h}\left(\cdot\right)$ is realized by a channel-wise convolutional layer with $\mathsf{C}\left(3, 1, 1\right)$.
The $\mathsf{Encode_k}\left(\cdot\right)$ contains two residual blocks which are same to the blocks in ResNet18 \cite{c25}.
The $\mathsf{Decode_k}\left(\cdot\right)$ is a four-layer upsampling network with the size of $\mathsf{D1}\left(4, 2, 16\right)$, $\mathsf{D2}\left(4, 2, 32\right)$, $\mathsf{D3}\left(4, 2, 64\right)$ and $\mathsf{C4}\left(3, 1, T_{pred}-T_{obs}\right)$.

In the prediction network, $\mathsf{f_c}$ has two convolutional layers, $\mathsf{C1}\left(3, 1, T_{pred}\right)$ and $\mathsf{C2}\left(3, 1, T_{pred}-T_{obs}\right)$. 
Besides, the two encoders and one decoder have the same configuration as those in the interaction network.

\subsubsection{Training Loss}
Our loss function involves two aspects, the image reconstruction loss and the interactive constraint loss.
Following common practice, we compute the mean square errors between the ground truth OGMs $\mathbb{O}_{gt}$ and the predicted $\mathbb{\widehat{O}}$ for all obstacles.
Mathematically, the reconstruction loss term is written as,
\begin{align*}
\mathcal{L}_r = & \frac{1}{NHWT} \sum_{i=1}^{N} \sum_{t=1}^{T} \mathds{1} \left[\left \|X_i^t\right \|_2 \leqslant \mathcal{R} \right] \left(\widehat{O}_{i}^{t}-O_{gt, i}^{t} \right)^{2} \tag{22} 
\end{align*}
where $\mathds{1} \left[\cdot\right]$ is an indicator function to remove the invalid objects which are out of the range $\mathcal{R}$.
Prediction time steps $T$ equals $T_{pred}-T_{obs}$ while $H$ and $W$ are the height and width of the OGM respectively.
In VisionNet, We set $\mathcal{R}=50,H=256,W=256$ for KITTI-Tracking dataset and $\mathcal{R}=10,H=128,W=128$ for UCY\&ETH datasets.

For a certain obstacle, the predicted positions generally locate on the safe regions with low energy in the ablated GDAS.
Thus, we  apply an L1 loss to improve the quality as well as the explanatory ability of the interactions,
\begin{align*}
\mathcal{L}_d = & \frac{1}{NHWT} \sum_{i=1}^{N}\sum_{t=1}^{T} \left \| O_{gt, i}^{t} \odot \left(M_c \odot  D_i^t\right) \right\|_1 \tag{23} 
\end{align*}

Overall, we minimize the weighted sum of the two losses,
\begin{align*}
\mathcal{L} = \alpha \mathcal{L}_r + \beta \mathcal{L}_d \tag{24} 
\end{align*}
where $\alpha$ and $\beta$ are weight parameters whose default value is 1.0 and 0.1 respectively.
Additionally, An Adam optimizer is employed to train the VisionNet for 30 epochs, with a learning rate of 0.0001. 

\section{EXPERIMENTS}
\label{section:4}

\begin{table*}[h]
	\caption{Comparison of Prediction Performance on ETH\&UCY Datasets}
	\label{table:1}
	\centering
	\small
	\renewcommand\arraystretch{1.2}
	\begin{tabular}{m{0pt}p{2cm}<{\centering}|p{1.6cm}<{\centering}|p{1.9cm}<{\centering}|p{2.0cm}<{\centering}|p{2.0cm}<{\centering}|p{2.8cm}<{\centering}|p{1.9cm}<{\centering}}
		\hline
		%\specialrule{0em}{1pt}{1pt}
		\rule{0pt}{12pt}&\normalsize Metric & \normalsize Dataset & \normalsize Linear &\normalsize Basic\ LSTM &\normalsize Social\ LSTM & \normalsize Social\ GAN(1v-1) & \normalsize $\bf{VisionNet}$ \\
		\hline
		\hline
		\rule{0pt}{0pt}&\multirow{5}{*}{\normalsize ADE} & \small ZARA-1 & 0.62 &\bf{0.41} & $ 0.47 $ & 0.42 & 0.60 \\
		%\cline{2-7}
		~ & ~ & \small ZARA-2 & 0.77 & 0.52 & 0.56 & 0.52 & $\bf{0.50}$ \\
		%\cline{2-7}
		~ & ~ & \small UNIV & 0.82 & 0.61 & 0.67 & 0.60 & $\bf{0.47}$\\
		%\cline{2-7}
		~ & ~ & \small ETH & 1.33 & 1.09 & 1.09 & 1.13 & $\bf{0.88}$\\
		%\cline{2-7}
		~ & ~ & \small HOTEL & \bf{0.39} & 0.86 & 0.79 & 1.01 & $ 0.69 $\\
		\hline
		\rule{0pt}{0pt}&\normalsize Average\ ADE & -  & 0.79 & 0.70 & 0.72 & 0.74 & $\bf{0.63}$\\
		\hline
		\hline
		\rule{0pt}{0pt}&\multirow{5}{*}{\normalsize FDE} & \small ZARA-1 & 1.21 & \bf{0.88} & 1.00 & 0.91  & 1.20 \\
		%\cline{2-7}
		~ & ~ & \small ZARA-2 & 1.48 & 1.11 & 1.17 & 1.11 & $\bf{1.02}$\\
		%\cline{2-7}
		~ & ~ & \small UNIV & 1.59 & 1.31 & 1.40 & 1.28 & $\bf{0.95}$\\
		%\cline{2-7}
		~ & ~ & \small ETH & 2.94 & 2.41 & 2.35 & 2.21 & $\bf{2.08}$ \\
		%\cline{2-7}
		~ & ~ & \small HOTEL & \bf{0.72} & 1.91 & 1.76 & 2.18 & 1.49 \\
		\hline
		\rule{0pt}{0pt}&\normalsize Average\ FDE & -  & 1.59 & 1.52 & 1.54 & 1.54 & $\bf{1.35}$ \\
		\hline
	\end{tabular}
\end{table*}

\subsection{Datasets and Metrics}

The VisionNet is evaluated on three public datasets: KITTI-Tracking, ETH and UCY.
The KITTI-Tracking contains raw LiDAR point clouds, images, localizations and object annotations.
The ETH and UCY datasets have 5 sets with a total of 1536 pedestrians in 4 crowded scenes and are annotated every 0.4 seconds (2.5Hz).

We demonstrate the effectiveness of the proposed interaction module as well as prediction performance on KITTI-Tracking dataset.
Additional ablation studies are conducted on various VisionNet architectures.
For extensive experiments, we compare the VisionNet with four determined algorithms on ETH and UCY datasets.
These methods include Linear Regression, Basic LSTM, Social LSTM and Social GAN (1v-1).
Different from stochastic models, these models output a single determined trajectory rather than $k$ random trajectories.
The VisionNet takes the past trajectory before $T_{obs}=8$  and forecasts the future trajectory for $T=12$ (until $T_{pred}=20$).
All experiments are conducted on a machine with an NVIDIA Tesla V100 GPU.

Similar to prior work \cite{c3,c13,c26}, three metrics are leveraged, 
\subsubsection{Mean Square Error (MSE)}
The mean square pixel-value difference between the predicted and the ground truth OGMs.
\subsubsection{Average Displacement Error (ADE)}
The mean Euclidean distance between predicted and the realistic trajectory.
\subsubsection{Final Displacement Error (FDE)}
The Euclidean distance between the final location of the predicted trajectory and the ground truth.

\begin{table}[h]
\caption{Ablation Study on KITTI-Tracking Dataset}
\label{table:2}
\centering
\small
\renewcommand\arraystretch{1.3}
\begin{tabular}{p{2.2cm}<{\centering}|p{0.4cm}<{\centering} p{0.4cm}<{\centering} p{0.4cm}<{\centering}|p{1.2cm}<{\centering}|p{1.2cm}<{\centering}}
	\hline
	\multirow{2}{*}{\normalsize Method} & \multicolumn{3}{c|}{\normalsize Component} & \multicolumn{2}{c}{\normalsize Metric} \\
	\cline{2-6}
	& \small BD &\small ND &\small IL &\small MSE &\small ADE \\
	\hline
	\hline
	\normalsize Baseline  & - & - & - & 11.07 & 2.17 \\
	\normalsize VisionNet-V1 & $\checkmark$ & - & - & 11.30 & 2.03\\
	\normalsize VisionNet-V2 & $\checkmark$ & $\checkmark$ & - &  \bf{10.76} & 2.01 \\
	\normalsize VisionNet-Full & $\checkmark$ & $\checkmark$ & $\checkmark$ & 11.18& \bf{1.71} \\
	\hline
\end{tabular}
\end{table}

\subsection{Experimental Results}
TABLE \ref{table:1} presents the performance of all determined models on ADE and FDE respectively.
While the Basic LSTM has the minimum error in ZARA-1 scene, the VisionNet, outperforms the others in ZARA-2, ETH and UNIV scenes.
On average, the VisionNet achieves significantly better performance, especially on the latter two crowded sets.
The improvements benefit from the reasonable representation and modeling for the interactions.
Previous approaches take only the spatial relations among obstacles into consideration.
However, relative attributes such as velocity, acceleration and orientation also affect the obstacles' intentions and result in collective interactions.
The proposed method takes full advantages of all these factors and builds the complex interactive relations through the B-DIS and the N-DIS, which is more comprehensive and reasonable.
Experiments on ETH and UCY datasets confirm the effectiveness of the VisionNet.

The ablation study of the VisionNet on KITTI-Tracking dataset is shown in TABLE \ref{table:2}.
BD and ND are denoted as basic driving spaces and noise driving spaces respectively.
IL is the interactive loss mentioned in Section \ref{section:3}.
Elaborated contrast experiments give evidence of the three advancements.
First, three versions of the VisionNets all achieve better performance than the baseline encoder-decoder architecture, which demonstrates the efficiency of the interaction modeling.
Considering the existence of the data noise, we introduce the noise driving spaces, which achieves further improvement in ADE.
Moreover, with the supervision of the additional interactive loss, our full version of the VisionNet acquires the best results.
Overall, the outcomes demonstrate the capacity of the VisionNet to model the collective interactions in complex scenarios.
Besides, it is worth mentioning that the VisionNet without interactive loss has the smallest MSE.
We credit the observation on a trade-off between image synthesis and interaction modeling.
Consequently, the balanced model provides the optimal solution for obstacle motion prediction.

\begin{figure*}[thpb]
	\centering
	\includegraphics[width=12.2cm]{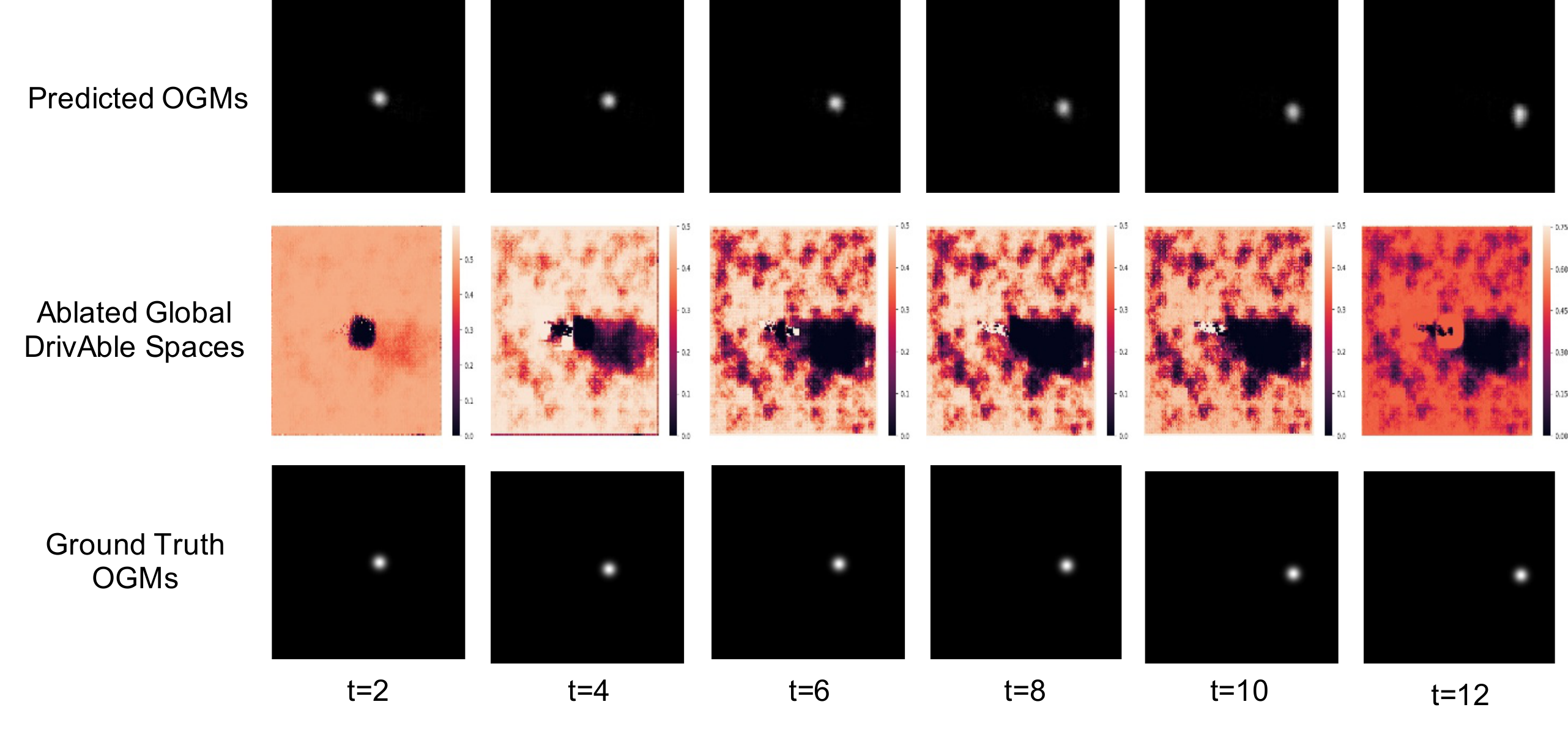}
	\caption{
		Qualitative results of our VisionNet. The top part is the predicted OGMs $\{\widehat{O}_j^t\}_{t=1}^{T_{obs}}$ of the VisionNet. The middle part is the ablated global drivable spaces $M_c \odot D$. The bottom part is the ground truth OGMs $\{{O_{gt}}_j^t\}_{t=T_{obs}+1}^{T_{pred}}$ of the predicted obstacle. Here, for clarity, the OGMs and the ablated GDAS are down-sampled by a factor of two in the time dimension.
	}
	\label{fig:3}
\end{figure*}

The visualization results of the VisionNet is shown in Figure \ref{fig:3}.
Three kinds of representations are displayed respectively, predicted OGMs, filtered global drivable spaces and ground truch OGMs.
It can be observed that the estimated positions of the VisionNet show high consistency with the ground truth locations.
When we filter out the global drivable spaces and remove the predicted agent, the heatmaps reveal low energy around the predicted positions, indicating the potential safe regions.
Besides, the visualization of the sequential drivable spaces verifies the common sense that the prediction turns to perform worse as the time elapses.

\section{CONCLUSION}
\label{section:5}
In this paper, we propose an vision-based motion prediction architecture, the VisionNet.
Consisting of two networks, it learns the collective interactions more comprehensibly through the basic and the noise driving spaces.
The interaction network generates the global drivable areas under the collective influence among dynamic on-road obstacles. 
The prediction network thus estimates the future trajectories more precisely with the interaction representation.
Different from the existing algorithms, our interaction modeling performs more explanatory and efficient.
Experiments on multiple public datasets strongly prove the effectiveness of the VisionNet and the results outperform other methods by a large margin.


\begin{thebibliography}{99}

\bibitem{c1} S. Srikanth, J. A.  Ansari, R. K. Ram, S. Sharma, J. K. Murthy and K. M. Krishna, ``INFER: INtermediate representations for FuturE pRediction," \textit{arXiv preprint} arXiv:1903.10641, 2019.
\bibitem{c2} N. Mohajerin and M. Rohani, ``Multi-step prediction of occupancy grid maps with recurrent neural networks," \textit{in 2019 IEEE Conference on Computer Vision and Pattern Recognition (CVPR)}. IEEE, 2019, pp. 10600-10608.
\bibitem{c3} J. Pan, C. Wang, X. Jia, J. Shao, L. Sheng, J. Yan and X. Wang, ``Video generation from single semantic label map," \textit{in 2019 IEEE Conference on Computer Vision and Pattern Recognition (CVPR)}. IEEE, 2019, pp. 3733-3742.
\bibitem{c4} Y. Hu, W. Zhan and M. Tomizuka, ``Probabilistic prediction of vehicle semantic intention and motion," \textit{in 2018 IEEE International Conference on Intelligent Vehicles Symposium (IV)}. IEEE, 2018, pp. 307-313.
\bibitem{c5} F. Altch$\acute{\rm e}$ and A. e. L. Fortelle, ``An LSTM network for highway trajectory prediction," \textit{in 2017 IEEE International Conference on Intelligent Transportation Systems (ITSC)}. IEEE, 2017, pp. 353-359.
\bibitem{c6} N. Deo and M. M. Trivedi, ``Multi-modal trajectory prediction of surrounding vehicles with maneuver based LSTMs," \textit{in 2018 IEEE International Conference on Intelligent Vehicles Symposium (IV)}. IEEE, 2018, pp. 1179-1184.
\bibitem{c7} C. Ju, Z. Wang, C. Long, X. Zhang, G. Cong, and D. E. Chang, ``Interaction-aware kalman neural networks for trajectory prediction," \textit{arXiv preprint} arXiv:1902.10928, 2019.
\bibitem{c8} H. M. Mandalia and D. D. Salvucci, ``Using support vector machines for lane-change detection," \textit{Proceedings of the human factors and ergonomics society annual meeting}, vol. 49, no. 22, pp. 1965-1969, 2005.
\bibitem{c9} J. Wiest, M. H$\ddot{\rm o}$ffken, U. Kreßel and K. Dietmayer, ``Probabilistic trajectory prediction with Gaussian mixture models," \textit{in 2012 IEEE International Conference on Intelligent Vehicles Symposium (IV)}. IEEE, 2012, pp. 141-146.
\bibitem{c10} H. Wu, Z. Chen, W. Sun, B. Zheng and W. Wang, ``Modeling trajectories with recurrent neural networks," \textit{in 2019 International Joint Conference on Artificial Intelligence (IJCAI)}. 2017, pp. 3083-3090.
\bibitem{c11} B. Kum, C. M. Kang, J. Kim, S. H. Lee, C. C. Chung and J. W. Choi, ``Probabilistic vehicle trajectory prediction over occupancy grid map via recurrent neural network," \textit{in 2017 IEEE International Conference on Intelligent Transportation Systems (ITSC)}. IEEE, 2017, pp. 339-404.
\bibitem{c12} A. Alahi, K. Goel, V. Ramanathan, A. Robicquet, F. Li and S. Savarese, ``Social LSTM: Human trajectory prediction in crowded spaces," \textit{in 2016 IEEE Conference on Computer Vision and Pattern Recognition (CVPR)}. IEEE 2016, pp. 961-971.
\bibitem{c13} Y. Zhu, D. Qian, D. Ren and H. Xia, ``StarNet: Pedestrian trajectory prediction using deep neural network in star topology," \textit{arXiv preprint} arXiv:1906.01797, 2019.
\bibitem{c14} A. Geiger, P. Lenz, and R. Urtasun, ``Are we ready for autonomous driving? The KITTI vision benchmark suite," \textit{in 2012 IEEE Conference on Computer Vision and Pattern Recognition (CVPR)}. IEEE 2012, pp. 3354-3361.
\bibitem{c15} S. Pellegrini, A. Ess, K. Schindler and L. V. Gool, ``You'll never walk alone: Modeling social behavior for multi-target tracking," \textit{in 2009 IEEE International Conference on Computer Vision (ICCV)}. IEEE, 2009, pp. 261-268.
\bibitem{c16} A. Lerner, Y. Chrysanthou and D. Lischinski, ``Crowds by example," \textit{Computer Graphics Forum}, vol. 26, no. 3, pp. 655-664, 2007.
\bibitem{c17} A. Gupta, J. Johnson, F. Li, S. Savarese and A. Alahi, ``Social GAN: Socially acceptable trajectories with generative adversarial networks," \textit{in 2018 IEEE Conference on Computer Vision and Pattern Recognition (CVPR)}. IEEE, 2018, pp. 2255-2264.
\bibitem{c18} A. Vemula, K. Muelling and J. Oh, ``Social attention: Modeling attention in human crowds," \textit{in 2018 IEEE International Conference on Robotics and Automation (ICRA)}. IEEE, 2018, pp. 1-7.
\bibitem{c19} J. Liang, L. Jiang, J. C. Niebles, A. Hauptmann and F. Li, ``Peeking into the future: Predicting future person activities and locations in videos," \textit{in 2019 IEEE Conference on Computer Vision and Pattern Recognition (CVPR)}. IEEE, 2019, pp. 5725-5734.
\bibitem{c20} Amir. Sadeghian, V. Kosaraju, Ali. Sadeghian, N. Hirose, S. H. Rezatofighi and S. Savarese, ``SoPhie: An attentive GAN for predicting paths compliant to social and physical constraints," \textit{in 2019 IEEE Conference on Computer Vision and Pattern Recognition (CVPR)}. IEEE, 2019, pp. 5725-5734.
\bibitem{c21} R. Chandra, U. Bhattacharya and A. Bera, ``TraPHic: Trajectory prediction in dense and heterogeneous traffic using weighted interactions," \textit{in 2019 IEEE Conference on Computer Vision and Pattern Recognition (CVPR)}. IEEE, 2019, pp. 8483-8492.
\bibitem{c22} N. Rhinehart, R. Mcallister, K. Kitani and S. Levine, ``PRECOG: Prediction conditioned on goals in visual multi-agent settings," \textit{arXiv preprint} arXiv:1905.01296, 2019.
\bibitem{c23} T. Wang, M. Liu, J. Zhu, G. Liu, A. Tao, J. Kautz and B. Catanzaro, ``Video-to-video synthesis," \textit{arXiv preprint} arXiv:1808.06601, 2018.
\bibitem{c24} P. Ondr$\acute{\rm u}$$\check{\rm s}$ka and I. Posner, ``Deep tracking: Seeing beyond seeing using recurrent neural networks," \textit{in 2016 Association for the Advancement of Artificial Intelligence (AAAI)}. 2016, pp. 3361-3367.
%\bibitem{c25} G. Welch and G. Bishop, ``An introduction to the kalman filter," \textit{in 2001 ACM Special Interest Group on Computer Graphics (SIGGRAPH)}, ACM, 2001, pp. 27599-23175.
\bibitem{c25} K. He, X, Zhuang, S. Ren and J. Sun, ``Deep residual learning for image recognition," \textit{in 2016 IEEE Conference on Computer Vision and Pattern Recognition (CVPR)}. IEEE, 2016, pp. 770-778.
\bibitem{c26} J. Amirian, J. Hayet and J. Pettre, ``Social Ways: Learning multi-modal distributions of pedestrian trajectories with GANs," \textit{arXiv preprint} arXiv:1808.06601, 2018.
\bibitem{c27} C. R. Qi, H. Su, K. Mo, and L. J. Guibas, ``Pointnet: Deep learning on point sets for 3d classification and segmentation," \textit{in 2017 IEEE Conference on Computer Vision and Pattern Recognition (CVPR)}. IEEE, 2017, pp. 652-660.

\end{thebibliography}
\end{document}